\documentclass{llncs}
\usepackage{makeidx} 
\begin{document}
\frontmatter          

\title{
The New AI: \\
General \& Sound \& Relevant for Physics \\ 
{\small Technical Report IDSIA-04-03, Version 2.0,
Nov 2003  (based on Version 1.0 \cite{Schmidhuber:03newai1}, Feb 2003).
{\em To appear in B. Goertzel and C. Pennachin, eds.:  },
{\em Artificial General Intelligence (accepted 2002)}} 
}

\newtheorem{postulate}{Postulate}

\def\odt{{\textstyle{1\over 2}}}
\def\maxarg{\mathop{\rm maxarg}}          
\def\minarg{\mathop{\rm minarg}}          

\date{}

\author{J\"{u}rgen Schmidhuber \\
{\tt juergen@idsia.ch - http://www.idsia.ch/\~{ }juergen}
}

\institute{IDSIA, Galleria 2, 6928 Manno (Lugano), Switzerland }

\maketitle

\begin{abstract}

Most traditional artificial intelligence (AI) systems of the past 
50 years are either very limited, or based on heuristics, or both. 
The new millennium, however, has brought substantial progress in 
the field of theoretically optimal and practically 
feasible algorithms for prediction, search, inductive
inference based on Occam's razor, problem solving, 
decision making, and reinforcement 
learning in environments of a very general type. 
Since inductive inference is at the heart of all inductive 
sciences, some of the results are relevant not only for AI and
computer science but also for physics, provoking nontraditional 
predictions based on Zuse's thesis of the 
computer-generated universe.

\end{abstract}


\section{Introduction}
\label{introduction}

Remarkably, there is a theoretically {\em optimal} way of
making predictions based on observations,  
rooted in the early work of Solomonoff and Kolmogorov
\cite{Solomonoff:64,Kolmogorov:65}.
The approach reflects basic principles of Occam's razor: simple
explanations of data are preferable to complex ones.

The theory of universal inductive inference quantifies what 
simplicity really means. 
Given certain very broad computability assumptions, it provides 
techniques for making optimally reliable statements about future events, 
given the past.

Once there is an optimal, formally describable
way of predicting the future, 
we should be able to construct a machine that 
continually computes and executes action sequences that maximize
expected or predicted reward, thus solving an ancient
goal of AI research.

For many decades, however, AI researchers 
have not paid a lot of attention to the theory of inductive
inference. Why not?  There is another reason
besides the fact that most of them have traditionally
ignored theoretical computer science:  the theory
has been perceived as being associated with excessive 
computational costs. In fact, its most general statements
refer to methods that are optimal (in  a certain asymptotic sense)
but incomputable.  
So researchers in machine learning and artificial 
intelligence have often resorted to alternative methods
that lack a strong theoretical foundation
but at least seem feasible in certain limited contexts.  
For example, since the early attempts 
at building a ``General Problem Solver''
\cite{Newell:63,SOAR:93}
much work has been done to develop
mostly heuristic machine learning
algorithms that solve new problems based on experience with
previous problems.  Many pointers to {\em learning by chunking,
learning by macros, hierarchical learning, learning by analogy,}
etc.  can be found in Mitchell's book \cite{Mitchell:97} and
Kaelbling's survey \cite{Kaelbling:96}.

Recent years, however, have brought substantial
progress in the field of {\em computable} and {\em feasible}
variants of optimal algorithms for prediction, search, inductive
inference, problem solving, decision making, and reinforcement 
learning in very general environments. In what follows I
will focus on the results obtained at IDSIA.

Sections 
\ref{shortest},
\ref{nonenumerable},
\ref{aixi}
relate Occam's razor and the notion of simplicity to the shortest 
algorithms for computing computable objects, and
will concentrate on recent {\em asymptotic} optimality results for
universal learning machines,
essentially ignoring issues of practical feasibility---compare 
Hutter's contribution \cite{Hutter:03aixigentle} in this volume.

Section \ref{speed}, however, will focus on our recent
non-traditional simplicity measure
which is {\em not} based on the shortest but on the {\em fastest} way
of describing objects, and
Section \ref{universe} will use this measure
to derive non-traditional
predictions concerning the future of our universe.

Sections
\ref{search},
\ref{oops},
\ref{oopsrl} 
will finally address quite pragmatic issues and ``true'' time-optimality: 
given a problem and only so much limited computation time,
what is the best way of spending it on evaluating 
solution candidates?
In particular, Section \ref{oops} will 
outline a bias-optimal way of incrementally solving
each task in a sequence of tasks with quickly verifiable 
solutions, given a probability distribution
(the {\em bias}) on programs computing solution candidates.
Bias shifts are computed by program prefixes that
modify the distribution on their suffixes by
reusing successful code for previous tasks
(stored in non-\-mo\-di\-fi\-able memory).
No tested program gets more
runtime than its probability times the total search time.
In illustrative experiments, ours becomes the first general  system to
{\em learn} a universal solver for arbitrary $n$ disk
{\em Towers of Hanoi} tasks (minimal solution size $2^n-1$).
It demonstrates the advantages of incremental learning by
profiting from previously solved, simpler tasks involving
samples of a simple context free language.
Sections \ref{oopsrl} discusses how to use this approach
for building general reinforcement learners. 

Finally,
Section \ref{gm} will summarize the recent 
G\"{o}del machine \cite{Schmidhuber:03gm2},
a self-referential, theoretically optimal self-improver
which explicitly addresses the
{\em `Grand Problem of Artificial Intelligence'}
\cite{Schmidhuber:03grandai}
by optimally dealing with limited resources in general reinforcement
learning settings.

\section{More Formally}
\label{formally}

What is the optimal way of predicting the future, given
the past? Which is the best way to act such as to maximize one's
future expected reward? Which is the best way of searching
for the solution to a novel problem, making optimal use of
solutions to earlier problems?

Most previous work
on these old and fundamental questions  has focused on
very limited settings, such as Markovian environments
where the optimal next action, given past inputs, depends
on the current input only \cite{Kaelbling:96}.

We will concentrate on a much weaker and therefore much more 
general assumption, namely, that the environment's responses are
sampled from a computable probability distribution.
If even this weak assumption were not true then we could not even formally
specify the environment, leave alone writing 
reasonable scientific papers about it.

Let us first introduce some notation.
$B^*$ denotes the set of finite
sequences over the binary alphabet $B=\{0,1\}$, $B^{\infty}$ the set of
infinite sequences over $B$, $\lambda$ the empty string, $B^{\sharp} =
B^* \cup B^{\infty}$.  $x,y,z,z^1,z^2$ stand for strings in $B^{\sharp}$.
If $x \in B^*$ then $xy$ is the concatenation of $x$ and $y$ (e.g.,
if $x=10000$ and $y=1111$ then $xy = 100001111$).  For $x \in B^*$,
$l(x)$ denotes the number of bits in $x$, where $l(x)= \infty$ for
$x \in B^{\infty}$; $l(\lambda) = 0$.  $x_{n}$ is the prefix of $x$
consisting of the first $n$ bits, if $l(x) \geq n$, and $x$ otherwise
($x_0 := \lambda$).  $\log$ denotes the logarithm with basis 2, $f,g$
denote functions mapping integers to integers.  We write $f(n)=O(g(n))$ if
there exist positive constants $c,n_0$ such that $f(n) \leq cg(n)$ for
all $n > n_0$.  For simplicity let us consider universal Turing Machines
\cite{Turing:36}
(TMs) with input alphabet $B$ and trinary output alphabet including the
symbols ``0'', ``1'', and `` '' (blank). For efficiency reasons,
the TMs should have several work tapes
to avoid potential quadratic slowdowns associated with 1-tape TMs.
The remainder of this paper assumes a fixed universal reference TM.

Now suppose bitstring $x$ represents the data observed so far.
What is its most likely continuation $y \in B^{\sharp}$? Bayes' theorem yields
\begin{equation}
P(xy \mid x)
= \frac{P(x \mid xy) P(xy)} 
       {P(x)}
\propto P(xy)
\label{bayes2}
\end{equation}
where $P(z^2 \mid z^1)$ is the probability of $z^2$, given knowledge
of $z^1$, and $P(x) = \int_{z \in B^{\sharp}}  P(xz) dz$ 
is just a normalizing factor.  So the most likely continuation
$y$ is determined by $P(xy)$, the {\em prior probability} of $xy$.
But which prior measure $P$ is plausible? 
Occam's razor suggests that the ``simplest'' $y$  should
be more probable.
But which exactly is the ``correct'' definition of simplicity?
Sections \ref{shortest} and \ref{nonenumerable}
will measure the simplicity of a description by its length.
Section \ref{speed} 
will measure the simplicity of a description by the time
required to compute the described object.

\section{Prediction Using a Universal 
Algorithmic Prior Based on the Shortest
Way of Describing Objects}
\label{shortest}

Roughly fourty years ago Solomonoff started the theory of
universal optimal induction based on the apparently harmless
simplicity assumption that $P$ is computable \cite{Solomonoff:64}.  
While Equation  (\ref{bayes2}) makes predictions of
the entire future, given the past,  
Solomonoff \cite{Solomonoff:78} focuses just on the next
bit in a sequence.  Although this provokes surprisingly nontrivial
problems associated with translating the bitwise approach to alphabets
other than the binary one --- this was achieved only recently
\cite{Hutter:01alpha+} --- it is sufficient for obtaining essential
insights.  Given an observed bitstring $x$, Solomonoff assumes the data are drawn
according to a recursive measure $\mu$; that is, there is a program for
a universal Turing machine that reads $x \in B^*$ and
computes $\mu(x)$ and halts. He estimates the probability of the next bit
(assuming there will be one), using the remarkable, well-studied, 
enumerable prior
$M$ \cite{Solomonoff:64,Zvonkin:70,Solomonoff:78,Gacs:83,LiVitanyi:97}
\begin{equation}
M(x) =  \sum_{program~prefix~p~computes \atop output~starting~with~x} 2^{-l(p)}.
\label{M}
\end{equation}
$M$ is {\em universal}, dominating the less general recursive 
measures as follows: For all $x \in B^*$,
\begin{equation}
M(x) \geq c_{\mu} \mu (x)
\end{equation}
where $c_{\mu}$ is a constant depending on $\mu$ but not on $x$.
Solomonoff observed that the conditional $M$-probability of a particular
continuation, given previous observations, converges towards the
unknown conditional $\mu$ as the observation size goes to infinity
\cite{Solomonoff:78}, and that the sum over all observation sizes of the
corresponding $\mu$-expected deviations is actually bounded by a constant.
Hutter (on the author's SNF research grant ``"Unification of Universal 
Induction and Sequential Decision Theory'')
recently showed that the number of prediction errors made by
universal Solomonoff prediction is essentially bounded by the number of
errors made by any other predictor, including the optimal scheme based
on the true $\mu$ \cite{Hutter:01alpha+}.  

{\bf Recent Loss Bounds for Universal Prediction.}
A more general recent result is this. Assume we do know that
$p$ is in some set $P$ of distributions. Choose a fixed weight $w_q$
for each $q$ in $P$ such that the $w_q$ add up to 1 (for simplicity,
let $P$ be countable). Then construct the Bayesmix
$M(x) = \sum_q w_q q(x)$, and predict using $M$ instead of the
optimal but unknown $p$.
How wrong is it to do that? The recent work of Hutter 
provides general and sharp (!) loss bounds \cite{Hutter:01loss+}:
   
Let $LM(n)$ and $Lp(n)$ be the total expected unit losses of the $M$-predictor
and the p-predictor, respectively, for the first $n$ events. Then
$LM(n)-Lp(n)$ is at most of the order of $\sqrt{Lp(n)}$. That is, $M$ is
not much worse than $p$. And in general, no other predictor can do
better than that!
In particular, if $p$ is deterministic, then the $M$-predictor soon
won't make any errors any more.
    
If $P$ contains {\em all} recursively computable distributions, then $M$
becomes the celebrated enumerable universal prior. That is, after
decades of somewhat stagnating research we now have sharp loss
bounds for Solomonoff's universal induction
scheme  (compare recent work of Merhav and Feder \cite{Feder:98}).

Solomonoff's approach, however, is uncomputable.
To obtain a feasible approach,
reduce M to what you get if you, say,  just add up weighted
estimated future finance data probabilities generated by 1000 commercial
stock-market prediction software packages. If only one of the probability
distributions happens to be close to the true one (but you do not know 
which) you still should get rich.

Note that the approach is much more general than what is normally done in
traditional statistical learning theory, e.g., \cite{Vapnik:95}, 
where the often quite unrealistic
assumption is that the observations are statistically independent.

\section{Super Omegas and Generalizations of Kolmogorov Complexity \& Algorithmic Probability} 
\label{nonenumerable}

Our recent research generalized Solomonoff's approach to the case of
less restrictive nonenumerable universal priors that are still computable
in the limit \cite{Schmidhuber:00v2,Schmidhuber:02ijfcs}.  

An object $X$ is formally describable if a finite amount of information
completely describes $X$ and only $X$.  More to the point, $X$ should be
representable by a possibly infinite bitstring $x$ such that there is a
finite, possibly never halting program $p$ that computes $x$ and nothing
but $x$ in a way that modifies each output bit at most finitely many
times; that is, each finite beginning of $x$ eventually {\em converges}
and ceases to change.  This constructive notion of formal describability
is less restrictive than the traditional notion of computability
\cite{Turing:36}, mainly because we do not insist on the existence
of a halting program that computes an upper bound of the convergence
time of $p$'s $n$-th output bit. Formal describability thus pushes
constructivism \cite{Brouwer:07,Beeson:85} to the extreme, barely
avoiding the nonconstructivism embodied by even less restrictive
concepts of describability (compare computability {\em in the limit}
\cite{Gold:65,Putnam:65,Freyvald:74} and $\Delta^0_n$-describability
\cite{Rogers:67}\cite[p. 46-47]{LiVitanyi:97}).

The traditional theory of inductive inference focuses on Turing machines 
with one-way write-only output tape. This leads to the universal enumerable
Solo\-mo\-noff\--Le\-vin (semi) measure. We introduced more general, nonenumerable,
but still limit-computable measures and
a natural hierarchy of generalizations of algorithmic
probability and Kolmogorov complexity \cite{Schmidhuber:00v2,Schmidhuber:02ijfcs},
suggesting that the ``true''
information content of some (possibly infinite) bitstring $x$ actually
is the size of
the shortest nonhalting program that converges to $x$ and nothing but $x$ on
a Turing machine that can edit its previous outputs. In fact,
this ``true'' content is often smaller than the traditional Kolmogorov complexity.
We showed that there are {\em Super Omegas} computable in the limit yet more
random than Chaitin's ``number of wisdom'' {\em Omega} \cite{Chaitin:87} (which
is maximally random in a weaker traditional sense),
and that any approximable
measure of $x$ is small for any $x$ lacking a short description.

We also showed that there is a
universal cumulatively enumerable measure of $x$ based on the measure of
all enumerable $y$ lexicographically greater than $x$. 
It is more dominant yet
just as limit-computable as Solomonoff's  \cite{Schmidhuber:02ijfcs}.
That is, if we are interested in limit-computable universal measures, 
we should prefer the novel universal cumulatively enumerable measure 
over the traditional enumerable one.
If we include in our Bayesmix such limit-computable distributions
we obtain again sharp loss bounds for prediction based on
the mix \cite{Schmidhuber:00v2,Schmidhuber:02ijfcs}.

Our approach highlights differences between countable
and uncountable sets.  Which are the potential consequences for physics?
We argue that things such as {\em un}countable time and space
and {\em in}computable probabilities actually should not play a role in
explaining the world, for lack of evidence that they 
are really necessary \cite{Schmidhuber:00v2}.
Some may feel tempted to counter this line of reasoning by pointing out
that for centuries physicists have calculated with continua of real
numbers, most of them incomputable.  Even quantum physicists who are
ready to give up the assumption of a continuous universe usually do
take for granted the existence of continuous probability distributions
on their discrete universes, and Stephen Hawking explicitly said: {\em
``Although there have been suggestions that space-time may have a
discrete structure I see no reason to abandon the continuum theories
that have been so successful.''} Note, however, that all physicists in
fact have only manipulated discrete symbols, thus generating finite,
describable proofs of their results derived from enumerable axioms.
That real numbers really {\em exist} in a way transcending the finite
symbol strings used by everybody may be a figment of imagination ---
compare Brouwer's constructive mathematics \cite{Brouwer:07,Beeson:85}
and the L\"{o}wenheim-Skolem Theorem \cite{Loewenheim:15,Skolem:19}
which implies that any first order theory with an uncountable model
such as the real numbers also has a countable model.  As Kronecker put
it: {\em ``Die ganze Zahl schuf der liebe Gott, alles \"{U}brige ist
Menschenwerk''} (``God created the integers, all else is the work of
man'' \cite{Cajori:19}).  Kronecker greeted with scepticism Cantor's
celebrated insight \cite{Cantor:1874} about real numbers,
mathematical objects Kronecker believed did not even exist.

Assuming our future lies among the few (countably many) describable
futures, we can ignore uncountably many nondescribable ones, in
particular, the random ones.
Adding the relatively mild assumption that the probability distribution
from which our universe is drawn is cumulatively enumerable provides
a theoretical justification of the prediction that the most likely
continuations of our universes are computable through short enumeration
procedures. In this sense Occam's razor is just a natural by-product 
of a computability assumption! 
But what about falsifiability?  
The pseudorandomness of our universe 
might be effectively undetectable in principle,
because some approximable and enumerable  patterns cannot be proven to
be nonrandom in recursively bounded time. 

The next sections, however, will introduce additional
plausible assumptions that do lead to {\em computable} optimal
prediction procedures.

\section{Computable Predictions through the Speed Prior Based on the 
Fastest Way of Describing Objects}
\label{speed}

Unfortunately, while $M$ and the more general priors of 
Section \ref{nonenumerable}
are computable in
the limit, they are not recursive, and thus
practically infeasible.  This drawback inspired less general yet
practically more feasible principles of minimum description length
(MDL) \cite{Wallace:68,Rissanen:86} as well as priors derived from
time-bounded restrictions \cite{LiVitanyi:97}
of Kolmogorov complexity 
\cite{Kolmogorov:65,Solomonoff:64,Chaitin:87}. 
No particular instance of these approaches, however, is universally
accepted or has a general convincing motivation that carries beyond rather
specialized application scenarios.  For instance, typical efficient MDL
approaches require the specification of a class of computable models of
the data, say, certain types of neural networks, plus some computable
loss function expressing the coding costs of the data relative to the
model. This provokes numerous {\em ad-hoc} choices.

Our recent work \cite{Schmidhuber:02colt}, however,
offers an alternative to the celebrated
but noncomputable algorithmic simplicity measure  
or Solomonoff-Levin measure discussed above
\cite{Solomonoff:64,Zvonkin:70,Solomonoff:78}.  
We introduced a new measure (a prior on the computable objects)
which is not based
on the {\bf shortest} but on the {\bf fastest} way of describing objects.

Let us assume that the observed data sequence
is generated by a computational process, and that any
possible sequence of observations is therefore computable in the limit
\cite{Schmidhuber:00v2}.
This assumption is stronger and more radical than the traditional one:
So\-lo\-mo\-noff just insists that the probability of any sequence prefix is
recursively computable, but the (infinite) sequence itself may still 
be generated probabilistically.

Given our starting
assumption that data are deterministically generated by a machine, it seems
plausible that the machine suffers from a computational resource
problem. Since some things are much harder to compute than others,
the resource-oriented point of view suggests the following postulate.

\begin{postulate}
\label{post}
The cumulative prior probability measure of all $x$ 
incomputable within time $t$ by any method
is at most inversely proportional to $t$.
\end{postulate}
This postulate leads to the Speed Prior $S(x)$, 
the probability that the output of the following 
probabilistic algorithm starts with $x$ \cite{Schmidhuber:02colt}:
\begin{quote}
{\bf Initialize:}
Set $t:=1.$ Let the input scanning head 
of a universal TM  point to the first
cell of its initially empty input tape.

{\bf Forever repeat:} 
While the number of instructions executed so far exceeds $t$: toss
an unbiased coin; if heads is up set $t:=2t$; otherwise exit. If the input
scanning head points to a cell
that already contains a bit, execute the corresponding instruction
(of the growing self-delimiting program, e.g., \cite{Levin:74,LiVitanyi:97}).
Else toss the coin again, set the cell's bit
to 1 if heads is up (0 otherwise), and set $t:=t/2.$

\end{quote}

Algorithm {\bf GUESS} is very similar to a probabilistic search
algorithm used in previous work on applied inductive inference
\cite{Schmidhuber:95kol,Schmidhuber:97nn}. 
On several toy problems it generalized
extremely well in a way unmatchable by traditional neural network
learning algorithms.

With $S$ comes a computable method {\bf AS} for predicting
optimally within $\epsilon$ accuracy \cite{Schmidhuber:02colt}.
Consider a finite but unknown program $p$ computing $y \in B^{\infty}$.
What if Postulate \ref{post} holds but $p$ is not optimally
efficient, and/or computed on a computer that differs from
our reference machine?  Then we effectively do not sample 
beginnings $y_k$ from $S$ but from an alternative semimeasure $S'$.
Can we still predict well? Yes, because the
Speed Prior $S$ dominates $S'$. 
This dominance is all we need to apply the recent loss bounds
\cite{Hutter:01loss+}.
The loss that we are expected to receive
by predicting according to {\bf AS}
instead of using the true but unknown $S'$ does not exceed 
the optimal loss by much \cite{Schmidhuber:02colt}.

\section{Speed Prior-Based Predictions for Our Universe}
\label{universe}

\begin{small}
\hspace{7.5cm}
{\sl ``In the beginning was the code.''} 

\hspace{3.5cm}
 {\sc First sentence of the Great Programmer's Bible}
 \end{small}
\vspace{0.3cm}

Physicists and economists
and other inductive scientists make predictions 
based on observations. 
Astonishingly, however, few physicists are aware of 
the theory of {\em optimal} inductive 
inference \cite{Solomonoff:64,Kolmogorov:65}.
In fact, when talking about the very
nature of their inductive business, many physicists cite 
rather vague concepts such as Popper's falsifiability
\cite{Popper:34}, instead of referring to quantitative results.

All widely accepted physical theories, however,
are accepted not because they
are falsifiable---they are not---or because they match the data---many
alternative theories also match the data---but because they are simple in
a certain sense.
For example, the theory of gravitation is induced from locally 
observable training examples such as falling apples and movements 
of distant light sources, presumably stars. The theory predicts 
that apples on distant planets in other galaxies
will fall as well. Currently nobody 
is able to verify or falsify this. But everybody believes in it 
because this generalization step makes the theory simpler than 
alternative theories with separate laws for apples on other planets.
The same holds for superstring theory \cite{Green:87} 
or Everett's many world theory \cite{Everett:57},
which presently also are neither verifiable nor falsifiable, yet 
offer comparatively simple explanations of numerous observations.
In particular, most of Everett's postulated many worlds will 
remain unobservable forever, 
but the assumption of their existence simplifies the theory, 
thus making it more beautiful and acceptable.

In Sections \ref{shortest} and \ref{nonenumerable} we have made
the assumption that the probabilities of next events, given previous
events, are (limit-)computable. Here we make a stronger assumption
by adopting {\bf Zuse's thesis} \cite{Zuse:67,Zuse:69}, namely,
that the very universe is actually being computed deterministically, 
e.g., on a cellular automaton (CA) \cite{Ulam:50,Neumann:66}.
Quantum physics, quantum computation \cite{Bennett:00,Deutsch:97,Penrose:89},
Heisenberg's uncertainty principle and
Bell's inequality \cite{Bell:66} do {\bf not} imply any physical 
evidence against this possibility, e.g.,  \cite{Hooft:99}. 

But then which is our universe's precise algorithm? 
The following method \cite{Schmidhuber:97brauer} does compute it:
\begin{quote}
Systematically create and execute all programs for a 
universal computer, such as a Turing machine or a CA; the first 
program is run for one instruction every second step on average, 
the next for one instruction every second of the
remaining steps on average, and so on.
\end{quote}
This method in a certain sense implements the simplest theory of 
everything: {\em all} computable universes, including ours and ourselves
as observers, are computed 
by the very short program that generates and executes {\em all} 
possible programs \cite{Schmidhuber:97brauer}. In nested fashion,
some of these programs will execute processes that again 
compute all possible universes, etc. \cite{Schmidhuber:97brauer}.
Of course, observers in ``higher-level'' universes may be completely 
unaware of observers or universes computed by nested processes, 
and vice versa.
For example, it seems hard to track and interpret the computations
performed by a cup of tea.

The simple method above is more efficient than it may seem at 
first glance. A bit of thought shows that it even has
the optimal order of complexity. For example, it
outputs our universe history as quickly as this
history's fastest program, save for a (possibly
huge) constant slowdown factor that does not 
depend on output size. 

Nevertheless, some universes are fundamentally harder to 
compute than others. This is reflected by 
the Speed Prior $S$ discussed above (Section \ref{speed}).
So let us assume that our universe's history is
sampled from $S$ or a less dominant prior reflecting
suboptimal computation of the history. 
Now we can immediately predict:

{\bf 1.}
Our universe will not get many times older than it is now
\cite{Schmidhuber:00v2} --- essentially,
the probability that it will last $2^n$ times longer than it has lasted
so far is at most $2^{-n}$.

{\bf 2.}
Any apparent randomness
in any physical observation
must be due to some yet unknown but
{\em fast} pseudo-random generator PRG \cite{Schmidhuber:00v2} which we
should try to discover.
{\bf 2a.}
A re-examination of beta decay patterns may reveal that
a very simple, fast, but maybe not quite trivial
PRG is responsible for the apparently random decays of neutrons
into protons, electrons and antineutrinos.
{\bf 2b.}
Whenever there are several possible continuations of our
universe corresponding to different Schr\"{o}dinger wave function
collapses --- compare Everett's widely accepted many worlds hypothesis
\cite{Everett:57} --- we should be more likely to end up in one computable
by a short {\em and} fast algorithm.  A re-examination of split experiment
data involving entangled states such as the observations of spins of
initially close but soon distant particles with correlated spins might
reveal unexpected, nonobvious, nonlocal algorithmic regularity due to
a fast PRG.

{\bf 3.}
Large scale quantum computation \cite{Bennett:00} will not
work well, essentially because it would require too many
exponentially growing computational
resources in interfering ``parallel universes'' \cite{Everett:57}.

{\bf 4.}
Any probabilistic algorithm depending on truly
random inputs from the environment  will not scale well in practice.

Prediction {\bf 2} is verifiable
but not necessarily falsifiable within a fixed time interval given in advance.
Still, perhaps the main reason for the
current absence of empirical evidence in this vein is that few \cite{Erber:85} 
have looked for it.

In recent decades several well-known physicists have started writing about 
topics of computer science, e.g., \cite{Penrose:89,Deutsch:97}, sometimes
suggesting that real world physics might allow for computing things that
are not computable traditionally. Unimpressed by this trend,
computer scientists have argued in favor of the opposite: since there is no
evidence that we need more than traditional computability to explain
the world, we should try to make do without this assumption, e.g.,
\cite{Zuse:67,Zuse:69,Fredkin:82,Schmidhuber:97brauer}.

\section{Optimal Rational Decision Makers}
\label{aixi}

So far we have talked about passive prediction, given the
observations. Note, however, that agents interacting with an environment
can also use predictions of the future to compute action sequences
that maximize expected future reward. Hutter's recent {\em AIXI model}
\cite{Hutter:01aixi+} 
(author's SNF grant 61847) 
does exactly this, by combining Solomonoff's
$M$-based universal prediction scheme with an {\em expectimax}
computation.  

In cycle $t$ action $y_t$ results in perception
$x_t$ and reward $r_t$, where all quantities may depend on
the complete history. The perception $x_t'$ and reward $r_t$ are sampled
from the (reactive) environmental probability distribution $\mu$. 
Sequential decision theory shows
how to maximize the total expected reward, called value,
if $\mu$ is known. Reinforcement learning \cite{Kaelbling:96} is used if $\mu$ is
unknown. AIXI defines a mixture distribution $\xi$
as a weighted sum of distributions $\nu \in \cal M$, where $\cal M$ is any class
of distributions including the true environment $\mu$. 

It can be shown that the conditional $M$ probability of 
environmental inputs to an AIXI agent, given the agent's earlier 
inputs and actions,
converges with increasing length of interaction against the true, unknown
probability \cite{Hutter:01aixi+},
as long as the latter is recursively computable, analogously
to the passive prediction case.

Recent work \cite{Hutter:02selfopt+} also demonstrated
AIXI's optimality in the following sense.  The
Bayes-optimal policy $p^\xi$ based on the mixture $\xi$ is self-optimizing
in the sense that the average value converges asymptotically for
all $\mu \in \cal M$ to the optimal value achieved by the (infeasible)
Bayes-optimal policy $p^\mu$ which knows $\mu$ in advance. 
The necessary condition that $\cal M$ admits self-optimizing policies 
is also sufficient. No other structural assumptions are made on $\cal M$. 
Furthermore, $p^\xi$ is Pareto-optimal in the sense that there is no other policy
yielding higher or equal value in {\em all} environments $\nu \in \cal M$
and a strictly higher value in at least one \cite{Hutter:02selfopt+}.

We can modify the AIXI model such that its predictions are based on the
$\epsilon$-approximable Speed Prior $S$ instead of the incomputable $M$. 
Thus we obtain
the so-called {\em AIS model.} Using Hutter's approach \cite{Hutter:01aixi+} 
we can now show that the conditional $S$
probability of environmental inputs to an AIS agent, given the earlier
inputs and actions, converges to the true but unknown probability,
as long as the latter is dominated by $S$, such as the $S'$ above.

\section{Optimal Universal Search Algorithms}
\label{search}

In a sense, searching is less general than reinforcement learning
because it does not necessarily involve predictions of unseen data.
Still, search is a central aspect of computer science (and any
reinforcement learner needs a searcher as a submodule---see 
Sections \ref{oopsrl} and \ref{gm}).  Surprisingly, however,
many books on search algorithms do not even mention the following, very
simple asymptotically optimal, ``universal'' algorithm for a broad class of
search problems. 

Define a probability distribution $P$ on a finite or infinite
set of programs for a given computer.  $P$ represents the searcher's 
initial bias (e.g., $P$ could be based on program length, or on a 
probabilistic syntax diagram).
\begin{quote}
Method {\sc Lsearch}:
\label{lsearch}
Set current time limit T=1.  {\sc While} problem not solved {\sc do:}
\begin{quote}
Test all programs $q$ such that $t(q)$,
the maximal time spent on creating and running and testing $q$,
satisfies $t(q) < P(q)~T$.  Set $T := 2 T.$
\end{quote}
\end{quote}
{\sc Lsearch} (for {\em Levin Search}) may be the algorithm
Levin was referring to
in his 2 page paper \cite{Levin:73} which states
that there is an asymptotically optimal universal search method for
problems with easily verifiable solutions, that is, solutions
whose validity can be quickly tested.
Given some problem class,
if some unknown optimal program $p$ requires $f(k)$ steps to solve a
problem instance of size $k$,
then {\sc Lsearch}
will need at most $O(f(k) / P(p)) = O(f(k))$ steps ---
the constant factor  $1/P(p)$ may be huge but does not depend on $k$.
Compare \cite[p. 502-505]{LiVitanyi:97} and \cite{Hutter:01fast+}
and the fastest way of computing all
computable universes in Section \ref{universe}.

Recently Hutter 
developed a more complex asymptotically
optimal search algorithm for {\em all} well-defined problems, not
just those with with easily verifiable solutions \cite{Hutter:01fast+}.
{\sc Hsearch} cleverly allocates part of the total
search time for searching the space of proofs to find provably correct
candidate programs with provable upper runtime bounds, and
at any given time
focuses resources on those programs with the currently
best proven time bounds.
Unexpectedly, {\sc Hsearch} manages to reduce
the unknown constant slowdown factor of {\sc Lsearch} to
a value of $1 + \epsilon$, where $\epsilon$ is an
arbitrary positive constant.
  
Unfortunately, however,  the search in proof space
introduces an unknown {\em additive} problem class-specific
constant slowdown, which again may be huge.
While additive constants generally are preferrable over
multiplicative ones, both types may make universal
search methods practically infeasible.

{\sc Hsearch}  and {\sc Lsearch} are nonincremental in the sense that
they do not attempt to minimize their constants by exploiting
experience collected in previous searches.
Our method
{\em Adaptive} {\sc Lsearch} or {\sc Als}
tries to overcome this
\cite{Schmidhuber:97bias} --- compare Solomonoff's related ideas
\cite{Solomonoff:86,Solomonoff:89}.
Essentially it works as follows: whenever {\sc Lsearch}
finds a program $q$ that computes a solution for the current problem,
$q$'s probability $P(q)$ is
substantially increased using a ``learning rate,''
while probabilities of alternative programs decrease
appropriately.
Subsequent  {\sc Lsearch}es for new problems then use the adjusted
$P$, etc.  A nonuniversal variant of this approach
was able to solve reinforcement learning (RL) tasks
\cite{Kaelbling:96}
in partially observable environments
unsolvable by traditional RL
algorithms \cite{Wiering:96levin,Schmidhuber:97bias}.
 
Each  {\sc Lsearch} invoked by {\sc Als} is optimal with respect to
the most recent adjustment of $P$.
On the other hand, the modifications of $P$ themselves
are not necessarily optimal.  Recent work discussed in the next section
overcomes this drawback in a principled way.

\section{Optimal  Ordered Problem Solver (OOPS)}
\label{oops}

Our recent
{\sc Oops} \cite{Schmidhuber:02oopsmlj,Schmidhuber:03nips} is a 
simple, general, theoretically sound, in a certain
sense time-optimal way of searching for a 
universal behavior or program that solves each problem 
in a sequence of computational problems,
continually organizing and managing and reusing earlier acquired knowledge.
For example, the $n$-th problem may be to compute the $n$-th event from 
previous events (prediction), 
or to find a faster way through a maze than the one found during the search 
for a solution to the  $n-1$-th problem (optimization).

Let us first introduce the important concept of bias-optimality,
which is a pragmatic definition of time-optimality, as opposed to
the asymptotic optimality of 
both {\sc Lsearch} and {\sc Hsearch}, which may be viewed
as academic exercises demonstrating that the $O()$ notation
can sometimes be practically irrelevant despite its
wide use in theoretical computer science.
Unlike asymptotic optimality, bias-optimality does not 
ignore huge constant slowdowns:

\begin{definition} [{\sc Bias-Optimal Searchers}]
\label{bias-optimal}
Given is a problem class $\cal R$,
a search space $\cal C$ of solution candidates
(where  any problem $r \in \cal R$ should have a solution in $\cal C$),
a task dependent bias in form of conditional probability
distributions $P(q \mid r)$ on the candidates $q \in \cal C$,
and a predefined procedure that creates and tests any given $q$
on any $r \in \cal R$ within time $t(q,r)$ (typically unknown in advance).
A searcher is {\em $n$-bias-optimal} ($n \geq 1$) if
for any maximal total search time $T_{max} > 0$
it is guaranteed to solve any problem $r \in \cal R$
if it has a solution $p \in \cal C$
satisfying $t(p,r) \leq P(p \mid r)~T_{max}/n$.
It is {\em bias-optimal} if  $n=1$.
\end{definition}
This definition makes intuitive sense: the most probable candidates
should get the lion's share of the total search time, in a way that
precisely reflects the initial bias.
Now we are ready to provide a general overview of the basic 
ingredients of {\sc oops} \cite{Schmidhuber:02oopsmlj,Schmidhuber:03nips}:

\noindent
{\bf Primitives.}
We start with an initial set of user-defined primitive behaviors. 
Primitives may be assembler-like instructions or
time-consuming software, such as, say, theorem provers, 
or matrix operators for neural network-like parallel architectures,  
or trajectory generators for robot simulations, 
or state update procedures for multiagent systems,
etc.  Each primitive is represented by a token.
It is essential that those primitives 
whose runtimes are not known in advance
can be interrupted at any time.

\noindent
{\bf Task-specific prefix codes.}
Complex behaviors are represented by token sequences or programs.
To solve a given task represented by task-specific program inputs,
{\sc oops} tries to sequentially compose
an appropriate complex behavior from primitive ones, 
always obeying the rules of a given user-defined
initial programming language. Programs are grown
incrementally, token by token; their beginnings 
or {\em prefixes} are immediately executed while being created;
this may modify some task-specific internal state 
or memory, and may transfer control back to previously 
selected tokens (e.g., loops).  
To add a new token to some program 
prefix, we first have to wait until the
execution of the prefix so far {\em explicitly
requests} such a prolongation, by setting an 
appropriate signal in the internal state.
Prefixes that cease to request any further tokens
are called {\em self-delimiting} programs or simply
programs  (programs are their own prefixes).
{\em Binary} self-delimiting programs were studied
by \cite{Levin:74} and \cite{Chaitin:75} in the context of
Turing machines \cite{Turing:36} and the
theory of Kolmogorov complexity and
algorithmic probability \cite{Solomonoff:64,Kolmogorov:65}.
{\sc Oops}, however, uses a more practical, not necessarily binary framework.

The program construction procedure above yields {\em task-specific
prefix codes} on program space: with any given task,
programs that halt because they have found a solution
or encountered some error cannot request any more tokens. 
Given the current task-specific inputs, no program can be the prefix of 
another one.  On a different task, however, the same program
may continue to request additional tokens. This is important
for our novel approach---incrementally growing self-delimiting 
programs are unnecessary for the asymptotic optimality
properties of {\sc Lsearch} and {\sc Hsearch}, but essential 
for {\sc oops}.

\noindent
{\bf Access to previous solutions.}
Let $p^n$ denote a found prefix solving the first $n$ tasks. 
The search for $p^{n+1}$ may greatly profit from
the information conveyed by (or the knowledge embodied by)
$p^1, p^2, \ldots, p^n$ which
are stored or {\em frozen} in special {\em non}modifiable 
memory shared by all tasks,  
such that they are accessible to $p^{n+1}$ (this is another difference to  
{\em non}incremental {\sc Lsearch} and {\sc Hsearch}).  
For example, $p^{n+1}$ might execute a token sequence that 
calls $p^{n-3}$ as a subprogram, or that 
copies $p^{n-17}$ into some internal {\em modifiable} task-specific memory, then modifies 
the copy a bit, then applies the slightly edited copy to the current task.
In fact, since the number of frozen programs may grow to a large value,
much of the knowledge 
embodied by $p^j$ may be about how to access 
and edit and use older $p^i$ ($i<j$).

\noindent
{\bf Bias.}
The searcher's initial bias is embodied by initial, 
user-defined, task dependent probability 
distributions on the finite or infinite search space of 
possible program prefixes. In the simplest case we start
with a maximum entropy distribution on the tokens, and
define prefix probabilities as the products of the 
probabilities of their tokens.
But prefix continuation probabilities may 
also depend on previous tokens in 
context sensitive fashion.

\noindent
{\bf Self-computed suffix probabilities.}
In fact, we permit that any executed prefix 
assigns a task-dependent, self-computed
probability distribution to its own possible continuations. 
This distribution
is encoded and manipulated in task-specific internal memory.
So unlike with {\sc Als} \cite{Schmidhuber:97bias} we do not use a prewired 
learning scheme to update the probability distribution.
Instead we leave such updates to prefixes
whose online execution modifies
the probabilities of their suffixes.
By, say,  invoking previously frozen code
that redefines the probability distribution on future prefix
continuations, the currently tested prefix may completely reshape the
most likely paths through the search space of its own continuations,
based on experience ignored by {\em non}incremental {\sc Lsearch} and {\sc Hsearch}.
This may introduce significant problem class-specific knowledge 
derived from solutions to earlier tasks. 

\noindent
{\bf Two searches.} 
Essentially, 
{\sc oops} provides equal resources for two near-{\em bias-optimal} 
searches 
(Def. \ref{bias-optimal})
that run in parallel until $p^{n+1}$
is discovered and stored in non-modifiable memory.  
The first is exhaustive; it systematically tests 
all possible prefixes on all tasks up to $n+1$.  
Alternative prefixes are tested 
on all current tasks in parallel while still growing; 
once a task is solved, we remove it from the current set;  
prefixes that fail on a single task are discarded.
The second search is much more focused; 
it only searches for prefixes that start with $p^n$, and 
only tests them on task $n+1$, which is safe, 
because we already know that such prefixes solve all tasks up to $n$.  

\noindent
{\bf Bias-optimal backtracking}.
{\sc Hsearch} and {\sc Lsearch}
assume potentially infinite storage. Hence they
may largely ignore questions of storage management.
In any practical system, however, we have to efficiently reuse
limited storage. Therefore, in both searches of {\sc oops}, alternative
prefix continuations are evaluated by a novel, practical, token-oriented
backtracking procedure 
that can deal with several tasks in parallel,
given some {\em code bias} in the form of previously found code.
The procedure always ensures near-{\em bias-optimality}
(Def. \ref{bias-optimal}):
no candidate behavior gets more time than it
deserves, given the probabilistic bias. 
Essentially we conduct a depth-first search in program space,
where the branches of the search tree are program prefixes,
and backtracking (partial resets of partially solved task sets and
modifications of internal states and continuation probabilities) 
is triggered once the sum of the runtimes of the current prefix on all current 
tasks exceeds the prefix probability multiplied by the total search time so far.

In case of unknown, infinite task sequences we can typically 
never know whether we already have found an optimal solver for all 
tasks in the sequence. But once we unwittingly do find one, 
at most half of the total future run time will be wasted on searching
for alternatives.  
Given the initial bias and subsequent
bias shifts due to $p^1, p^2, \ldots, $ no other bias-optimal
searcher 
can expect to solve the $n+1$-th task set
substantially faster than {\sc oops}.  A by-product of this
optimality property is that it gives us 
a natural and precise measure of bias and bias shifts,
conceptually related to Solomonoff's {\em conceptual jump size} 
of \cite{Solomonoff:86,Solomonoff:89}.

Since there is no fundamental difference between
domain-specific problem-solving programs and programs
that manipulate probability distributions and thus essentially rewrite
the search procedure itself, we collapse both learning and
metalearning in the same time-optimal framework.

\noindent
{\bf An example initial language.} 
For an illustrative application, we wrote an interpreter for a 
stack-based universal programming language inspired
by {\sc Forth} \cite{Forth:70},
with initial primitives for defining and calling recursive
functions, iterative loops, arithmetic operations, and domain-specific
behavior.
Optimal metasearching for better search algorithms is enabled
through the inclusion of bias-shifting instructions that can modify the
conditional probabilities of future search options in currently running
program prefixes. 

\noindent
{\bf Experiments.} 
Using the assembler-like language mentioned above,
we first teach {\sc oops} something about recursion, by training it to
construct samples of the simple context free language $\{ 1^k2^k \}$ 
($k$ 1's followed by $k$ 2's),
for $k$ up to 30 (in fact, the system discovers a 
universal solver for all $k$).
This takes roughly 0.3 days on a standard personal computer (PC).
Thereafter, within a few additional days, 
{\sc oops} demonstrates incremental knowledge transfer:
it exploits aspects of its previously 
discovered universal $1^k2^k$-solver,
by rewriting its search procedure such that it
more readily discovers a universal solver
for all $k$ disk {\em Towers of Hanoi} problems---in
the experiments it solves all
instances up to $k=30$ (solution size $2^k-1$),
but it would also work for $k>30$.
Previous, less general reinforcement learners
and {\em non}learning AI planners 
tend to fail for much smaller instances.

\noindent
{\bf Future research}
may focus on devising particularly compact,
particularly reasonable sets of initial codes with
particularly broad practical applicability.
It may turn out that the most useful initial
languages are not traditional programming
languages similar to the {\sc Forth}-like
one, but instead based on a handful of primitive 
instructions for massively parallel cellular automata
\cite{Ulam:50,Neumann:66,Zuse:69},
or on a few nonlinear operations on matrix-like
data structures such as those used in
recurrent neural network research
\cite{Werbos:74,Rumelhart:86,Bishop:95}.
For example, we could use the principles of
{\sc oops} to create a non-gradient-based,  near-bias-optimal
variant of Hochreiter's successful recurrent network
metalearner \cite{Hochreiter:01meta}.
It should also be of interest to study probabilistic
{\em Speed Prior}-based {\sc oops} variants \cite{Schmidhuber:02colt}
and to devise applications of {\sc oops}-like methods as
components of universal reinforcement learners  (see below).
In ongoing work, we are applying {\sc oops} to the problem
of  optimal trajectory planning for robotics
in a realistic physics simulation.
This involves the interesting trade-off
between comparatively fast program-composing primitives or
{\em ``thinking primitives''}
and time-consuming {\em ``action primitives''},
such as {\em stretch-arm-until-touch-sensor-input}.

\section{OOPS-Based Reinforcement Learning}
\label{oopsrl}

At any given time, a reinforcement learner \cite{Kaelbling:96} will 
try to find a {\em policy} (a strategy for future decision
making) that maximizes its expected future reward.
In many traditional reinforcement learning (RL) applications, 
the policy that works best in a given set of training trials
will also be optimal in future test trials \cite{Schmidhuber:01direct}.
Sometimes, however, it won't.  To see the difference between 
searching (the topic of the previous sections) and 
reinforcement learning (RL), consider an agent and two boxes.
In the $n$-th trial the agent may open and collect the content of
exactly one box. The left box will contain
$100n$ Swiss Francs, the right box $2^n$ Swiss Francs, 
but the agent does not know this
in advance. During the first 9 trials
the optimal policy is {\em ``open left box.''} This is what a good
searcher should find, given the outcomes of the first 9 trials.
But this policy will be suboptimal in trial 10.
A good reinforcement learner, however, should extract the
underlying regularity in the reward generation process
and predict the future tasks and rewards, picking the right box in 
trial 10, without having seen it yet.

The first general, asymptotically optimal reinforcement learner is
the recent AIXI model \cite{Hutter:01aixi+,Hutter:02selfopt+} (Section \ref{aixi}).
It is valid for a very broad class of
environments whose reactions to action sequences (control signals)
are sampled from arbitrary computable probability distributions.
This means that AIXI is far more general than traditional RL approaches. 
However, while AIXI clarifies the theoretical limits of
RL, it is not practically feasible, just like {\sc Hsearch} is not.  
From a pragmatic point of view,
what we are really interested in is a reinforcement learner
that makes optimal use of given, limited computational resources.
In what follows, we will outline one way of using {\sc oops}-like bias-optimal
methods as components of general yet feasible reinforcement learners.

We need two {\sc oops} modules. The first is
called the predictor or world model. The second is
an action searcher using the world model.  The life of the entire system 
should consist of a sequence of {\em cycles} 1, 2, ...
At each cycle, a limited amount of computation time will
be available to each module.
For simplicity we assume that during each cyle the system
may take exactly one action.
Generalizations to actions consuming several cycles are straight-forward though.
At any given cycle, the system executes the following procedure:

\begin{enumerate}
\item
For a time interval fixed in advance,
the predictor is first trained
in bias-optimal fashion to find a better world model,
that is, a program that predicts the inputs from the environment
(including the rewards, if there are any), given a history of 
previous observations and actions.
So the $n$-th task ($n=1,2,\ldots$)
of the first {\sc oops} module is to find (if possible) a 
better predictor than the best found so far. 

\item
Once the current cycle's time for predictor improvement is used up,
the current world model (prediction program) found by
the first {\sc oops} module will be used by
the second module, again in bias-optimal fashion,
to search for a future action sequence that maximizes 
the predicted cumulative reward (up to some time limit). 
That is, the $n$-th task ($n=1,2,\ldots$)
of the second {\sc oops} module will be to find 
a control program that computes a control sequence of actions,
to be fed into the program representing the current world model 
(whose input predictions are successively fed back to itself in
the obvious manner),
such that this control sequence leads to higher predicted
reward than the one generated by the best control program 
found so far.

\item
Once the current cycle's time for control program search 
is used up, we will execute the current action of the best control 
program found in step 2. Now we are ready for the next cycle.

\end{enumerate}
The approach is reminiscent of an earlier, heuristic, non-bias-optimal
RL approach based on two adaptive recurrent neural networks, 
one representing the world model, the other one a controller
that uses the world model to extract a policy for maximizing
expected reward \cite{Schmidhuber:91nips}. The method was inspired by
previous combinations of {\em non}recurrent, 
{\em reactive} world models and controllers 
\cite{Werbos:87specifications,NguyenWidrow:89,JordanRumelhart:90}.

At any  given time, until which temporal horizon should the predictor
try to predict?  In the AIXI case, the proper way of treating the temporal horizon
is not to discount it exponentially, as done in most traditional
work on reinforcement learning, but to let the future horizon grow
in proportion to the learner's lifetime so far \cite{Hutter:02selfopt+}.
It remains to be seen whether this insight carries over to {\sc oops-rl}. 

Despite the bias-optimality properties of {\sc oops} for
certain ordered task sequences, however, {\sc oops-rl}
is not necessarily the best way of spending limited time in general
reinforcement learning situations. On the other hand,
it is possible to use {\sc oops} as
a proof-searching submodule of the recent, optimal, universal,
reinforcement learning G\"{o}del machine \cite{Schmidhuber:03gm2}
discussed in the next section.

\section{The G\"{o}del Machine}
\label{gm}

The G\"{o}del machine \cite{Schmidhuber:03gm2}
explicitly addresses the {\em `Grand Problem of Artificial Intelligence'}
 \cite{Schmidhuber:03grandai}
 by optimally dealing with limited resources in general reinforcement
 learning settings,
 and with the possibly huge (but constant) slowdowns buried
 by AIXI$(t,l)$ \cite{Hutter:01aixi+}
 in the somewhat misleading $O()$-notation.
It is designed to solve arbitrary computational problems beyond those
solvable by plain {\sc oops},
such as maximizing the expected future reward of a robot in a possibly
stochastic and reactive environment
(note that the total utility of some robot behavior may be hard
to verify---its evaluation may consume the robot's entire lifetime).
 
 How does it work?
 While executing some arbitrary 
 initial problem solving strategy, the G\"{o}del machine
 simultaneously runs a proof searcher which systematically and repeatedly
 tests proof techniques.  Proof techniques are programs that may read any
 part of the G\"{o}del machine's state, and write on a reserved part which may be reset
 for each new proof technique test.  In an example G\"{o}del machine
 \cite{Schmidhuber:03gm2} this writable
 storage includes the variables {\em proof} and {\em switchprog}, where
 {\em switchprog} holds a potentially unrestricted program whose execution
 could completely rewrite any part of the G\"{o}del machine's current software.
 Normally the current {\em switchprog} is not executed.  However, proof
 techniques may invoke a special subroutine {\em check()} which tests
 whether {\em proof} currently holds a proof showing that the utility of
 stopping the systematic proof searcher and transferring control to the
 current {\em switchprog} at a particular
 point in the near future exceeds the
 utility of continuing the search until some alternative {\em switchprog}
 is found.  Such proofs are derivable from the proof searcher's axiom
 scheme which formally describes the utility function to be maximized
 (typically the expected future reward in the expected remaining lifetime
 of the G\"{o}del machine), the computational costs of hardware instructions (from 
 which all programs are composed), and the effects of hardware instructions on
 the G\"{o}del machine's state.  The axiom scheme also formalizes known probabilistic
 properties of the possibly reactive environment, and also the {\em
 initial} G\"{o}del machine state and software, which includes the axiom scheme itself
 (no circular argument here).  Thus proof techniques can reason about
 expected costs and results of all programs including the proof searcher.
  
  Once {\em check()} has identified a provably
  good {\em switchprog}, the latter is executed
  (some care has to be taken here because the proof
  verification itself and the transfer of control to {\em switchprog}
  also consume part of the typically limited lifetime).
  The discovered {\em switchprog} represents a {\em globally}
  optimal self-change in the following sense:
  provably {\em none} of all the alternative {\em switchprog}s
  and {\em proof}s (that could be found in the future
  by continuing the proof search) is worth
  waiting for.

There are many ways of initializing the
proof searcher.  Although identical proof techniques may
yield different proofs depending on the time
of their invocation (due to the continually
changing G\"{o}del machine state), there is a bias-optimal and
asymptotically optimal proof searcher initialization
based on a variant of {\sc oops} \cite{Schmidhuber:03gm2} (Section \ref{oops}).
It exploits the fact that proof verification is a simple and
fast business   where the particular optimality
notion of {\sc oops} is appropriate.
 The G\"{o}del machine itself, however, may have
 an arbitrary, {\em typically different and more powerful} sense
 of optimality embodied by its given utility function.

\section{Conclusion}

Recent theoretical and practical advances are currently driving a
renaissance in the fields of universal learners 
and optimal search \cite{Schmidhuber:02nipsws}.  
A new kind of AI is emerging. 
Does it really deserve the attribute {\em ``new,''} given that 
its roots date back to the 1930s, when 
G\"{o}del published the fundamental
result of theoretical computer science \cite{Goedel:31}
and Zuse started to build the first general purpose computer 
(completed in 1941), and the 1960s, when Solomonoff and
Kolmogorov published their first relevant results?
An affirmative answer seems justified, since
it is the recent results on practically feasible computable
variants of the old incomputable methods 
that are currently reinvigorating the long dormant field.
The ``new'' AI is new in the sense that it abandons the mostly
heuristic or non-general approaches of the past decades,
offering methods that are both general and theoretically 
sound, and provably optimal in a sense that {\em does} 
make sense in the real world.

We are led to claim that the future will belong to universal
or near-universal learners that are 
more general than traditional reinforcement learners / decision makers
depending on strong Markovian assumptions, or than learners based on
traditional statistical learning theory, which often require unrealistic
i.i.d. or Gaussian assumptions. Due to ongoing hardware advances
the time has come for optimal search in algorithm space, as opposed
to the limited space of reactive mappings embodied by traditional methods
such as artificial feedforward neural networks.

It seems safe to bet that not only computer scientists but also
physicists and other inductive scientists will start to pay more
attention to the fields of universal induction and optimal search, since
their basic concepts are irresistibly powerful and general and simple.
How long will it take for these ideas to unfold their full impact? A very
naive and speculative guess driven by wishful thinking might be based
on identifying the {\em ``greatest moments in computing history''} and
extrapolating from there.  Which are those ``greatest moments''?  Obvious
candidates are: 
\begin{enumerate}
\item
{\em 1623:} first mechanical calculator by Schickard
starts the computing age (followed by
machines of Pascal, 1640, and Leibniz, 1670).
\item
{\em Roughly two centuries later:} 
concept of a {\em programmable} computer (Babbage, UK, 1834-1840).
\item
{\em One century later:}
fundamental theoretical work on universal integer-based programming
languages and the limits of proof and computation (G\"{o}del, Austria,
1931, reformulated by Turing, UK, 1936); 
first working programmable computer (Zuse, Berlin, 1941).

{\small \em (The next 50 years saw many theoretical
advances as well as faster and faster switches---relays were replaced by
tubes by single transistors by numerous transistors 
etched on chips---but arguably this was rather predictable, incremental 
progress without radical shake-up events.)}
\item
{\em Half a century later:}
World Wide Web (UK's Berners-Lee, Switzerland, 1990).  
\end{enumerate}
This list seems to
suggest that each major breakthrough tends to come roughly twice as fast as the
previous one. Extrapolating the trend, optimists should expect the next radical
change to manifest itself one quarter of a century after the most recent one, that is,
by 2015, which happens to coincide with the date when the fastest
computers will match brains in terms of raw computing power, according to
frequent estimates based on Moore's law.  The author is confident that
the coming 2015 upheaval (if any) will involve universal learning algorithms
and G\"{o}del machine-like, optimal, incremental search in algorithm space
\cite{Schmidhuber:03gm2}---possibly laying
a foundation for the remaining series of faster and faster additional
revolutions culminating in an ``Omega point'' expected around 2040.

\section{Acknowledgments}
Hutter's frequently mentioned work was funded through the
author's SNF grant 2000-061847
``Unification of universal inductive inference and
sequential decision theory.''
Over the past three decades,
numerous discussions with Christof Schmidhuber 
(a theoretical physicist) helped to crystallize the ideas
on computable universes---compare his notion 
of {\em ``mathscape''} \cite{Christof:00}.

\bibliography{bib}
\bibliographystyle{plain}
\end{document}